\documentclass[runningheads]{llncs}
\usepackage[T1]{fontenc}
\RequirePackage{amsmath}
\usepackage{graphicx}
\usepackage{amssymb}
\usepackage{multirow}
\usepackage[utf8]{inputenc}
\usepackage{url}
\usepackage{booktabs}
\usepackage{xspace}
\usepackage{microtype}
\usepackage[hidelinks]{hyperref}
\usepackage[normalem]{ulem}
\useunder{\uline}{\ul}{}
\usepackage{pifont}
\usepackage[dvipsnames,svgnames]{xcolor}         
\usepackage{colortbl}
\def\eg{\emph{e}.\emph{g}.}

\usepackage[capitalize]{cleveref}
\crefname{section}{Sec.}{Secs.}
\Crefname{section}{Section}{Sections}
\Crefname{table}{Table}{Tables}
\crefname{table}{Tab.}{Tabs.}

\newcommand{\ci}[1]{\footnotesize{~($\pm #1$)}}
\newcommand{\Ours}{MammoFlow}

\begin{document}

\title{MammoFlow: Multiview Mammogram Synthesis with Anatomically Consistent Flow Matching}
\titlerunning{MammoFlow}

\author{
    Yuexi Du\inst{1} \and
Leya Barrientos\inst{2} \and
Laura Sheiman\inst{2} \and
John Lewin\inst{2} \and \\ 
Hemant D. Tagare\inst{1,2} \and 
Nicha C. Dvornek \inst{1,2} 
}

\authorrunning{Y. Du et al.}

\institute{Department of Biomedical Engineering, Yale University,
\and
Department of Radiology \& Biomedical Imaging, Yale University,
\\
\email{\{yuexi.du, nicha.chitphakdithai\}@yale.edu}}

\maketitle

\begin{abstract}
Multiview mammography relies on paired craniocaudal (CC) and mediolateral oblique (MLO) views to provide complementary projections of a 3D breast volume, enabling precise anomaly localization. However, acquiring high-quality, balanced datasets remains challenging for deep learning applications. We propose a novel method to synthesize multiview mammograms by leveraging the inherent geometric relationship between CC and MLO views. To enforce an implicit 3D consistency prior during generation, we develop an alignment module that searches a 2D affine transformation subspace to establish optimal anatomical correspondence. Leveraging this alignment, we introduce a pixel-space self-consistency loss based on the Earth Mover's Distance (EMD) between the 1D anteroposterior (AP) axis tissue distributions of the generated images. Integrated into a pretrained flow matching model, \Ours~forces synthesized pairs to share physically plausible tissue distributions from the chest wall to the nipple. To our knowledge, this is the first work to guide multiview mammogram generation using implicit geometric tissue correspondence. Our method demonstrates superior image quality, passes expert radiologist evaluation, and generates physically consistent pairs that improve downstream classification AUC by 5\%.
\footnote{Code and pretrained model: \url{https://github.com/XYPB/MammoFlow}.}
\keywords{Multiview Mammography \and Image Synthesis \and Anatomic Consistency}
\end{abstract}

\section{Introduction}

\begin{figure}[!t]
    \centering
    \includegraphics[width=\columnwidth]{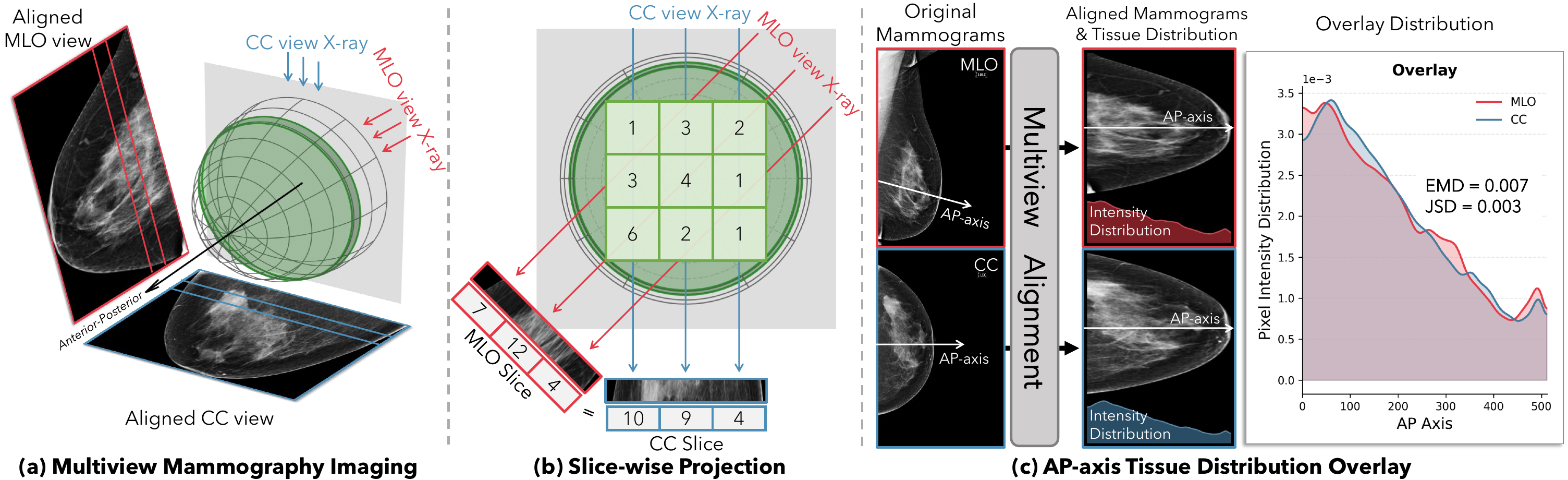}
    \caption{{\bf Multiview Mammography} (a) Multiview mammograms project a shared 3D volume. (b) Total tissue intensity is conserved across corresponding anteroposterior (AP) slices. (c) Aligned AP-axis distributions demonstrate high cross-view correlation.}
    \label{fig:teaser}
\end{figure}

Breast cancer remains a leading cause of cancer-related mortality among women worldwide \cite{sung2021global}. As the gold standard for screening, full-field digital mammography (FFDM) is widely adopted due to its accessibility and sensitivity. During screening, radiologists rely on paired craniocaudal (CC) and mediolateral oblique (MLO) views to precisely localize anomalies, benefiting from the complementary perspectives these projections provide (\cref{fig:teaser}(a)). This dual-view foundation is also critical for modern computer-aided diagnosis systems~\cite{wang2023dual,yamazaki2022two,petrini2022breast,ghosh2024mammo,du2025geometry}. While data-driven deep learning methods have achieved remarkable success in FFDM analysis~\cite{chang2025artificial}, acquiring high-quality, paired datasets remains hindered by privacy concerns and annotation costs, and the number of cancer cases comprises a small fraction of the curated samples. Consequently, limited and highly imbalanced data restrict the further development of deep learning applications in mammography.

Standard data augmentation, such as affine transformations and tumor artifacting~\cite{walsh2022comparison}, is commonly applied to alleviate data scarcity~\cite{sorkhei2021csaw}. 
However, these augmentations operate based on GT images and cannot introduce true morphological diversity. 
Generative image synthesis has thus emerged as a promising alternative to create plausible mammograms for training downstream models~\cite{heng2024survey}. Yet, existing approaches~\cite{montoya2024mam,garrucho2023high} primarily focus on single-view generation, neglecting the inherent multiview nature of clinical practice. While view-to-view translation frameworks like CA3D-Diff~\cite{li2025bidirectional} address this gap, they still require a ground-truth reference view. A recent study, Mammo-RGB~\cite{garza2025mammorgb}, attempts simultaneous dual-view generation by stacking views in color channels; however, it ignores explicit 3D anatomical relationships, which often result in cross-view artifacts.

In this work, we propose \Ours, a novel multiview mammogram synthesis method guided by the explicit anatomical correlation between CC and MLO views. As illustrated in \cref{fig:teaser}(b), the 2D projection along the anteroposterior (AP) axis in both views originates from the same 3D breast volume; thus, the sum of their tissue intensities along this axis should be largely conserved. Inspired by this projection prior, we introduce an alignment module that searches a subspace of 2D affine transformations to spatially align the CC and MLO projections (\cref{fig:teaser}(c)). Leveraging this alignment, we design a novel self-consistency loss based on the Earth Mover's Distance (EMD). By formulating EMD as a differentiable geometric constraint, we minimize the distributional divergence of tissue along the AP axis during training. Integrating this module into a large-scale rectified flow model~\cite{esser2024scaling} allows our method to generate high-fidelity, dual-view mammograms that maintain 3D anatomical consistency. Extensive quantitative analysis and expert radiologist evaluations demonstrate the superiority of our generated images, which further show a non-trivial improvement in downstream classification performance when utilized as additional training data.

\section{Methods}

\begin{figure}[!t]
    \centering
    \includegraphics[width=\columnwidth]{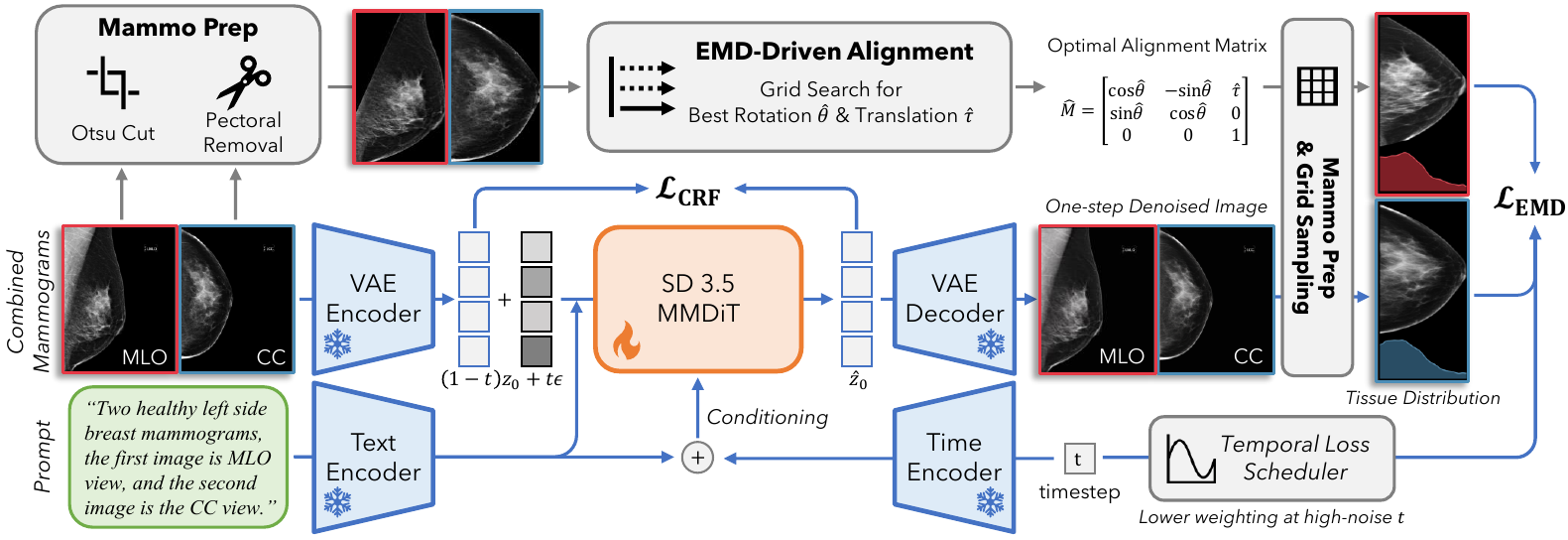}
    \caption{{\bf Proposed Multiview Mammogram Synthesis Pipeline} We propose to use the tissue correspondence relationship of the ground-truth images to guide the rectified flow (RF) training process. Besides the Conditional RF loss $\mathcal{L}_{CRF}$, we further optimize the Earth-Mover's Distance (EMD) loss over the denoised images. We use the Temporal Loss Scheduler to reduce the influence of a blurry reconstructed image during training.}
    \label{fig:method}
\end{figure}

We first introduce the Conditional Flow Matching (CFM) image synthesis backbone of \Ours, followed by the novel EMD-driven view alignment module and the multiview EMD constraint loss. An overview of \Ours~is in~\cref{fig:method}. During training, each raw CC/MLO pair and its text prompt are used for the CRF objective, and the real paired images provide the precomputed alignment used to evaluate $\mathcal{L}_{\text{EMD}}$ on the one-step reconstructed pair. During inference, the trained model takes only Gaussian noise and a prompt to generate novel paired views; no ground-truth image or alignment is required.

\subsection{Conditional Flow Matching}

Our multiview mammogram synthesis framework builds upon the conditional rectified flow (CRF) formulation~\cite{liu2022flow,lipman2022flow}, which constructs straight paths between the data distribution and a standard normal distribution. Given a combined multiview mammogram $x=[x_{MLO}, x_{CC}]$ and its corresponding text condition $y$ (\cref{fig:method}), we extract the clean latent representation $z_0$ using a pre-trained variational autoencoder (VAE)~\cite{kingma2013auto}. The linear probability path connecting $z_0$ to standard Gaussian noise $\epsilon \sim \mathcal{N}(0, I)$ is defined as:
\begin{equation}
    z_t = (1 - t)z_0 + t\epsilon, \quad t \in [0, 1]
\end{equation}
where $t$ represents the continuous timestep. The denoising model, parameterized as $\mathbf{v}_\theta(z_t, t, y)$, is trained to predict the velocity of this deterministic flow~\cite{liu2022flow,esser2024scaling}. Since the target velocity is $\frac{\partial z_t}{\partial t} = \epsilon - z_0$, the CRF objective minimizes the expected squared difference between the predicted velocity and this target trajectory:
\begin{equation}
    \mathcal{L}_{\text{CRF}} = \mathbb{E}_{z_0, \epsilon, t, y} \left[ \| \mathbf{v}_\theta(z_t, t, y) - (\epsilon - z_0) \|^2 \right]
\end{equation}
During inference, the model generates novel samples by solving the corresponding ordinary differential equation (ODE). Because the defined flow is linear, the predicted velocity directly points toward the clean data, allowing us to approximate the one-step denoised latent $\hat{z}_0$ at any timestep $t$:
\begin{equation}
    \hat{z}_0 = z_t - t \mathbf{v}_\theta(z_t, t, y)
\end{equation}

\subsection{EMD-Driven Alignment}

Unlike CC views, MLO projections are captured at an angle and routinely include the pectoral muscle to visualize the chest wall. Consequently, the anatomical anteroposterior (AP) axis in the MLO view is inherently tilted relative to the image grid. Furthermore, view-specific, non-rigid breast compression introduces additional morphological variations. These characteristics make the direct comparison of spatial intensity values across raw mammograms geometrically inconsistent. 

To resolve this, we propose an EMD-driven alignment module (\cref{fig:method}). We first preprocess the raw images by removing text labels via Otsu thresholding, cropping the image to the breast region. For the MLO view, we apply the Hough transform to detect the pectoral muscle boundary and mask this region with zero intensity to prevent it from skewing the tissue distribution.

Next, we align the two views such that their AP axes are parallel to the horizontal image axis. Based on standard imaging protocols, we can assume that the CC view's AP axis is already horizontally aligned~\cite{sweeney2018review}. Therefore, we restrict our geometric alignment to a 2D affine subspace consisting of a rotation angle $\theta$ and a horizontal translation $\tau$ applied to the MLO view.
Since we trimmed the excess region and resized the MLO image after the affine transform, the stretch term is also considered implicitly.
We denote this affine transformation as $\mathcal{A}(x_{MLO}; \theta, \tau)$. 
To evaluate the alignment quality, we measure the divergence between the 1D AP-axis tissue distributions. Let $\Phi(\cdot)$ denote the differentiable 1D projection operator, which sums the raw pixel intensities vertically at each position along the AP axis and normalizes by total image intensity. To account for non-rigid tissue deformation, $\Phi(\cdot)$ includes a horizontal 1D Gaussian smoothing filter, providing spatial relaxation. We utilize the Earth Mover's Distance (EMD) to quantify the tissue correlation, which is defined as:
\begin{equation}
    \text{EMD}(p, q) = \textstyle\sum_{i=1}^{W} |P[i] - Q[i]|
\end{equation}
where $P$ and $Q$ are the cumulative distribution functions of the normalized tissue distributions $p$ and $q$, respectively, and $W$ is the image width.
Unlike Jensen-Shannon divergence (JSD), 
EMD explicitly penalizes the physical displacement of tissue along the AP axis, ensuring continuous and stable training gradients.

Because there are only two degrees of freedom ($\theta, \tau$), we formulate the alignment as a discrete optimization problem, using grid search to find the optimal parameters $\hat{\theta}$ and $\hat{\tau}$ that minimize the distributional distance between the views:
\begin{equation}
    (\hat{\theta}, \hat{\tau}) = \underset{\theta, \tau}{\arg\min}~\text{EMD}\big(\Phi(x_{CC}), \Phi(\mathcal{A}(x_{MLO}; \theta, \tau))\big)
\end{equation}
By heuristically bounding the search ranges for $\theta$ and $\tau$ based on physical imaging protocols, we ensure the search space is highly compressed, resulting in a computationally lightweight alignment process.

\subsection{Multiview EMD Regularization}

The EMD-driven alignment module provides the optimal transformation parameters $(\hat{\theta}, \hat{\tau})$ for the ground-truth image pair. We leverage this geometric prior to guide the training of the rectified flow model. Because the physical tissue constraint must be evaluated in the image domain, we first decode the predicted clean latent $\hat{z}_0$ to obtain reconstructed dual-view mammograms $\hat{x} = [\hat{x}_{CC}, \hat{x}_{MLO}]$. 

To ensure the synthesized views maintain anatomically consistent 3D breast anatomy, we optimize the EMD between the generated views during training. Since the prediction $\hat{x}$ is spatially anchored to the ground-truth $x$ conditioning, we can directly apply the precomputed optimal alignment parameters $(\hat{\theta}, \hat{\tau})$ to the generated MLO view. Similarly, we reuse the precomputed Otsu thresholding mask and pectoral region mask so that this process is differentiable. We implement this affine transformation $\mathcal{A}(\cdot)$ using differentiable grid sampling, allowing the gradient to flow back through the VAE decoder. The self-consistency EMD loss for the generated mammogram pair is defined as:
\begin{equation}
    \mathcal{L}_{\text{EMD}} = \text{EMD}\big(\Phi(\hat{x}_{CC}), \Phi(\mathcal{A}(\hat{x}_{MLO}; \hat{\theta}, \hat{\tau}))\big)
\end{equation}

\noindent\textbf{Temporal Loss Scheduler.} 
During CRF training, the model predicts the clean latent $\hat{z}_0$ in a single step from timestep $t$. While the deterministic nature of flow matching ensures the velocity vector points toward the original data, one-step predictions from high-noise timesteps (large $t$) often lack structural fidelity. Because the EMD loss relies on anatomical geometry, applying it to highly blurry predictions is uninformative and can destabilize training. 
To mitigate the influence of poorly structured $\hat{x}$ at high noise levels, we introduce a temporal loss scheduler. We apply a cosine decay schedule based on the continuous timestep $t \in [0, 1]$ to suppress the EMD penalty at large $t$ and upweight it as $t$ approaches $0$. 
Our final optimization objective is thus formulated as:
\begin{equation}
    \mathcal{L} = \mathcal{L}_{\text{CRF}} + \tfrac{\lambda}{2}\big(\cos(\pi t) + 1\big) \mathcal{L}_{\text{EMD}},
\end{equation}
where $\lambda$ is the weight of $\mathcal{L}_{\text{EMD}}$. The $\mathcal{L}_{\text{EMD}}$ is effective only during training, making it a plug-and-play module for general flow-matching models. Moreover, our method brings no extra cost during inference.

\section{Experiments}
\label{sec:experiment}

\begin{table}[!t]
\centering
\caption{\textbf{Image Generation Results.} We report the FID and FrD for image quality and the relative difference (\%) of mean EMD and JSD from the ground truth values to measure multiview correlation. * indicates the method requires a ground truth image as a reference. Best synthetic results are in bold. Our method is shaded in gray.}
\label{tab:main}
\setlength{\tabcolsep}{3.5pt}
\resizebox{0.95\textwidth}{!}
{
\begin{tabular}{lcccccccccccc}
\toprule
\multicolumn{1}{c}{\multirow{2}{*}{\textbf{Method}}} & \multicolumn{4}{c}{\textbf{CSAW}} & \multicolumn{4}{c}{\textbf{VinDr}} & \multicolumn{4}{c}{\textbf{RSNA}} \\ 
\cmidrule(lr){2-5} \cmidrule(lr){6-9} \cmidrule(lr){10-13}
 & FID & FrD & $\Delta$EMD & $\Delta$JSD & FID & FrD & $\Delta$EMD & $\Delta$JSD & FID & FrD & $\Delta$EMD & $\Delta$JSD \\ \midrule \midrule
GT (Oracle) & 8.89 & 3.02 & - & - & 8.36 & 5.42 & - & - & 9.76 & 12.0 & - & - \\
Rand. Shuff. & - & - & 72.14 & 123.54 & - & - & 61.35 & 83.95 & - & - & 64.92 & 83.91 \\ \midrule
CA3D-Diff* & 63.3 & 8.72 & 14.12 & 16.89 & 70.5 & 6.72 & 173.83 & 498.86 & 64.5 & 8.73 & 37.20 & 90.98 \\ \midrule
Mammo-RGB & 88.6 & 14.9 & 60.67 & 188.53 & 102.8 & 21.5 & 33.01 & 49.57 & 148.6 & 15.1 & 15.43 & \textbf{20.98} \\
Vanilla (w/o $\mathcal{L}_{\text{EMD}})$ & 73.4 & 17.7 & 10.82 & 68.52 & 70.5 & 48.8 & 18.37 & 54.61 & 66.3 & 16.4 & 7.75 & 28.86 \\ 
\rowcolor[HTML]{EFEFEF} \Ours & \textbf{53.3} & \textbf{6.60} & \textbf{1.08} & \textbf{9.57} & \textbf{67.5} & \textbf{12.4} & \textbf{4.19} & \textbf{15.22} & \textbf{65.4} & \textbf{12.7} & \textbf{2.73} & 33.48 \\ \bottomrule

\end{tabular}
}

\end{table}
\begin{table}[!t]
\centering
\caption{\textbf{Ablation Study.} Performance of different multiview designs on CSAW. Best results are in bold; second best are underlined. Our method is shaded in gray.}
\label{tab:ablation}
\setlength{\tabcolsep}{5.5pt}
\resizebox{0.95\textwidth}{!}
{
\begin{tabular}{cccccccc}
\toprule
$\mathcal{L}_{\text{EMD}}$ & Temporal Loss Scheduler & Gaussian Smoothing & $\lambda$ & FID & FrD & $\Delta\text{EMD}$ & $\Delta\text{JSD}$ \\ \midrule\midrule
- & - & - & - & 60.41 & 6.95 & 10.82 & 68.52 \\
\checkmark & - & \checkmark & 0.10 & 64.71 & 14.26 & 18.39 & {\ul 15.06} \\
\checkmark & cosine & - & 0.10 & 74.67 & 51.72 & 10.82 & 193.54 \\
\checkmark & linear & \checkmark & 0.10 & {\ul 54.54} & \textbf{6.59} & 14.40 & 22.27 \\
\checkmark & cosine & \checkmark & 1.00 & 83.42 & 7.22 & {\ul 10.62} & 104.66 \\
\rowcolor[HTML]{EFEFEF} \checkmark & cosine & \checkmark & 0.10 & \textbf{53.32} & {\ul 6.60} & \textbf{1.08} & \textbf{9.57} \\
\checkmark & cosine & \checkmark & 0.01 & 66.93 & 12.52 & 24.80 & 358.77 \\ \bottomrule
\end{tabular}
}

\end{table}

\subsection{Experimental Settings} 

\noindent\textbf{Datasets.} We evaluate our method on three public mammography datasets: (1) \textbf{CSAW}~\cite{Strand2022CSAW-CC} contains 98k paired mammograms, including 740 cases with visible cancer based on segmentation masks (80\%/20\% train/test split). (2) \textbf{VinDr}~\cite{nguyen2023vindr} includes 20k paired mammograms with $\sim$1k cancer cases (official train/test split). (3) \textbf{RSNA}~\cite{rsna-breast-cancer-detection} provides 54k images and $\sim$1k cancer cases (85\%/15\% train/test split). All images are resized to $512 \times 512$. 

\noindent\textbf{Implementation.} We use a pre-trained Stable Diffusion 3.5 Medium~\cite{esser2024scaling} backbone. We freeze all components except the Multimodal Diffusion Transformer (MMDiT) at training. Multiview mammograms are spatially concatenated along the horizontal axis. We fine-tune the model in half-precision on a single NVIDIA H200 GPU using the AdamW optimizer, a constant learning rate of $1\times10^{-6}$, and a batch size of 16 for 40k steps. For the EMD-driven alignment, the search space for $\theta$ is $[-10^\circ, 10^\circ]$ with a $2^\circ$ step size and for $\tau$ is $[-64, 64]$ pixels with an 8-pixel step size. The initial $\mathcal{L}_{\text{EMD}}$ weight is set to $\lambda=0.1$ and applied with a cosine decay scheduler. Inference is performed using 100 steps, sampling 1k images for downstream evaluation. A fixed random seed of 42 is used throughout the experiment.
We train the downstream classifier with a learning rate of $1\times10^{-3}$, a weight decay of $0.1$, and a batch size of 32 for 25 epochs on each dataset.

\noindent\textbf{Baselines.} We compare our approach against \textbf{CA3D-Diff}~\cite{li2025bidirectional}, a view-to-view translation model requiring an oracle reference, and \textbf{Mammo-RGB}~\cite{garza2025mammorgb}, a simultaneous multiview generation method reproduced by ourselves with the same backbone. We also include a \textbf{Vanilla} baseline without the proposed $\mathcal{L}_{\text{EMD}}$.

\noindent\textbf{Metrics.} Models are evaluated on two aspects: (1) \textit{Image Quality}, measured via \textbf{FID}~\cite{Seitzer2020FID} for overall image quality and \textbf{FrD}~\cite{konz2026frd} for radiomic feature consistency relevant to diagnosis; (2) \textit{Multiview Correspondence}, evaluated as the relative percentage difference in mean divergence between the generated and the ground truth (GT) AP-axis tissue distribution: $\Delta D = 100\% \times \frac{|\text{mean}(D_{syn}) - \text{mean}(D_{GT})|}{\text{mean}(D_{GT})}$, where $D$ denotes either $\text{EMD}$ or $\text{JSD}$. We apply affine alignment only when computing EMD/JSD for paired images; all other evaluations use raw synthetic images.

\noindent\textbf{Downstream Tasks.} To evaluate practical utility, we fine-tune our model on malignant cases to synthesize cancerous data distributions, where the prompt controls the malignancy. We then train an EfficientNet-B2~\cite{tan2019efficientnet} multiview binary cancer classifier (concatenating view features into an MLP) by augmenting the real dataset with 1k and 5k synthetic malignant pairs. Additionally, two expert radiologists performed a blinded reader study on 80 multiview mammograms (20 pairs per method). For each pair, they assessed authenticity (real vs. synthetic) and anatomical pairing correctness across views.

\subsection{Results}

\begin{table}[t]
  \centering
  \begin{minipage}[t]{0.58\textwidth}
    \centering
    \caption{\textbf{Cancer Classification.} 
    We add different amounts of synthetic cancer images for training breast cancer classification and  report AUC-ROC on the original test set. Best results are in bold.
    }
    \label{tab:classification}
    \setlength{\tabcolsep}{8pt}
    \resizebox{\linewidth}{!}
    {
    \begin{tabular}{cccc}
    \toprule
    \textbf{Training Data} & \textbf{CSAW} & \textbf{VinDr} & \textbf{RSNA} \\ \midrule\midrule
    Real & .7452 & .7775 & .7676 \\
    Real + 1k Synth. & .7899 & .8055 & .7711 \\
    Real + 5k Synth. & \textbf{.7904} & \textbf{.8196} & \textbf{.7822} \\ \bottomrule
    \end{tabular}
    }
  \end{minipage}
  \hfill
  \begin{minipage}[t]{0.40\textwidth}
    \centering
    \caption{\textbf{Reader Study.} 
    We report ratio (mean $\pm$ standard deviation) of identifying the multiview mammograms as real and paired.
    }
    \label{tab:human_eval}
    \setlength{\tabcolsep}{3pt}
    \resizebox{\linewidth}{!}
    {
    \begin{tabular}{lcc}
    \toprule
    \multicolumn{1}{c}{\textbf{Method}} & \textbf{Authenticity} & \textbf{Pairing} \\ \midrule\midrule
    GT & 97.5\ci{2.5} & 100.0\ci{0.0} \\ \midrule
    CA3D-Diff & 2.5\ci{2.5} & 82.5\ci{2.5} \\
    Mammo-RGB & 0.0\ci{0.0} & 95.0\ci{5.0} \\
    \Ours & \textbf{37.5}\ci{12.5} & \textbf{97.5}\ci{2.5} \\ \bottomrule
    \end{tabular}
    }
  \end{minipage}
\end{table}

\noindent\textbf{Quantitative Evaluation.} Main results are in~\cref{tab:main}. Computing FID and FrD between two subsets of the GT test set establishes an oracle lower bound. To validate our multiview alignment metrics, we evaluate randomly shuffled GT pairs, observing a significant increase in both $\Delta$EMD and $\Delta$JSD ($p < 0.0001$, Mann-Whitney $U$ test), confirming these metrics accurately capture geometric misalignment. 
Our \Ours~consistently outperforms other multiview generation methods in both image quality and multiview correspondence. Notably, no significant difference was found comparing the EMD distribution of our synthetic pairs against the ground truth on the CSAW dataset ($p=0.3089$, Mann-Whitney $U$ test), confirming the high geometric fidelity of our generations. The addition of $\mathcal{L}_{\text{EMD}}$ not only improves cross-view tissue correlation but also improves overall radiomic fidelity compared to the vanilla baseline. While CA3D-Diff achieves slightly lower FrD scores on the VinDr and RSNA datasets, this is expected as it relies on an oracle ground truth reference image. However, CA3D-Diff still exhibits inferior multiview correspondence because it assumes a fixed $45^\circ$ rigid rotation between views, ignoring the patient-specific adjustments made in clinical practice. Mammo-RGB fails to match our alignment performance, as stacking views in color channels inherently ignores explicit spatial correlation.

\noindent\textbf{Ablation Study.} \cref{tab:ablation} highlights the contribution of each component. Unconstrained paired generation (w/o $\mathcal{L}_{\text{EMD}}$) degrades both structural anatomy and image quality. Applying the EMD loss with a constant weight distorts breast structures due to harmful gradient signals at high-noise timesteps. Removing the Gaussian spatial relaxation limits the ability to handle non-rigid tissue deformation. Among temporal schedulers, cosine decay performs best by mainly penalizing structural divergence only at low-noise timesteps. $\mathcal{L}_{EMD}$ weight of $\lambda=0.1$ achieves the best balance between visual fidelity and geometric alignment.

\begin{figure}[!t]
    \centering
    \includegraphics[width=0.95\columnwidth]{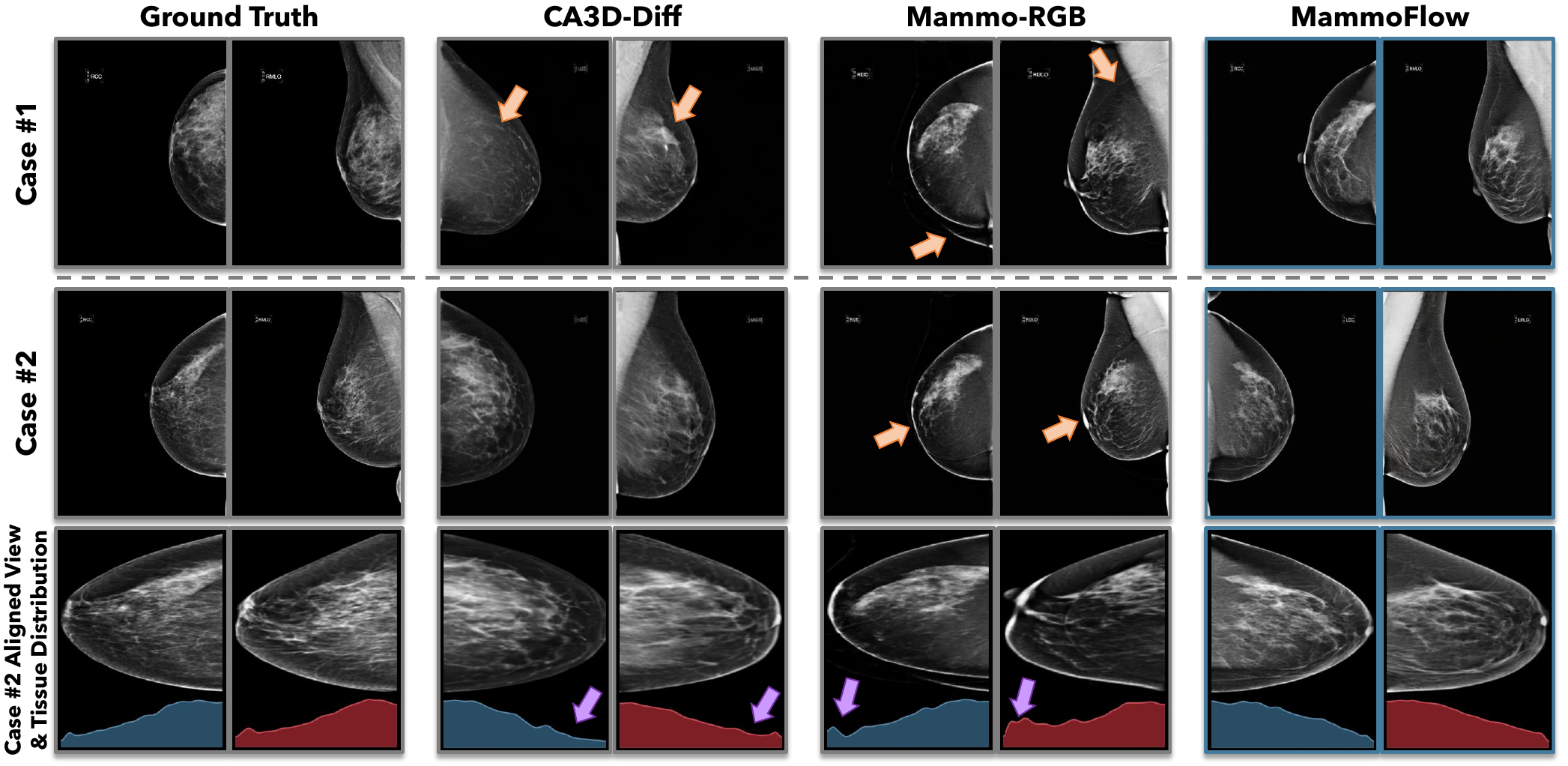}
    \caption{{\bf Qualitative Results.} We visualize two random CC/MLO pairs generated from Gaussian noise with the same prompt for each method. The first column shows GT images. Rows 1--2 show synthetic cases, and the final row plots the aligned view and AP-axis tissue distribution for case 2. The orange arrow highlights artifacts and mismatched anatomical structure. The purple arrow highlights mismatched tissue distribution.}
    \label{fig:visualize}
\end{figure}

\noindent\textbf{Downstream Classification.} As reported in~\cref{tab:classification}, incorporating our synthetic images improves the breast cancer classification AUC by up to 5\% on CSAW, demonstrating the high diagnostic value of our generated anatomies.

\noindent\textbf{Reader Study.} \cref{tab:human_eval} shows experts can easily identify GT images. Among the generative models, \Ours~was the most deceptive, with $37.5\%$ of images rated as real. Furthermore, our generated pairs achieved a $97.5\%$ pairing success rate, surpassing the oracle-guided CA3D-Diff framework.

\noindent\textbf{Qualitative Results.} Visual comparisons in~\cref{fig:visualize} confirm that our method demonstrates the most anatomically consistent multiview generation. Baselines frequently synthesize artifacts (\eg, contours of the other view in Mammo-RGB samples) or localized structures (\eg, high-density tissue) in one view that are absent in the other (orange arrows). The plotted AP-axis distributions further demonstrate this: while our synthesized distributions closely mirror the native mass distributions of the GT data, other methods show obvious tissue density mismatches along the projection axis (purple arrows).

\section{Discussion and Conclusion}

We presented \Ours, a multiview mammogram synthesis framework leveraging a flow-matching model and implicit 3D tissue guidance. By embedding an EMD-driven alignment module, our method minimizes cross-view spatial divergence, generating anatomically consistent pairs that significantly improve downstream classification.

Limitations include increased training GPU memory ($\sim20\%$) due to pixel-space constraints requiring VAE backpropagation.
Furthermore, our 2D affine alignment approximates complex 3D compressions, and classical pectoral masking struggles with very dense anatomies. Future work will optimize computational efficiency and model fully non-rigid 3D deformations across views.

\noindent\textbf{Acknowledgments} This work was supported by NIH grant R21EB032950.

\noindent\textbf{Disclosure of Interests} The authors have no competing interests in this work and other related research.

\bibliographystyle{splncs04}
\bibliography{ref}
\setcounter{tocdepth}{1}
\end{document}